\documentclass[sigconf]{acmart}
\usepackage{subcaption}
\usepackage{pifont}
\usepackage{multirow}
\usepackage[capitalize]{cleveref} 

\acmConference[Conference acronym 'XX]{Make sure to enter the correct
  conference title from your rights confirmation email}{June 03--05,
  2018}{Woodstock, NY}
\acmISBN{978-1-4503-XXXX-X/2018/06}

\acmSubmissionID{1916}

\begin{document}

\title{Mitigating Long-tail Distribution in Oracle Bone Inscriptions: Dataset, Model, and Benchmark}


\author{Jinhao Li}
\authornote{Both authors contributed equally to this research.}
\email{lomljhoax@stu.ecnu.edu.cn}
\affiliation{%
  \institution{East China Normal University}
  \city{Shanghai}
  \country{China}
}
\orcid{1234-5678-9012}

\author{Zijian Chen}
\authornotemark[1]
\email{zijian.chen@sjtu.edu.cn}
\affiliation{%
  \institution{Shanghai Jiao Tong University}
  \city{Shanghai}
  \country{China}
}

\author{Runze Jiang}
\email{20040329jrz@sjtu.edu.cn}
\affiliation{%
  \institution{Shanghai Jiao Tong University}
  \city{Shanghai}
  \country{China}
}

\author{Tingzhu Chen}
\authornote{Corresponding Authors.}
\email{tingzhuchen@sjtu.edu.cn}
\affiliation{%
  \institution{Shanghai Jiao Tong University}
  \city{Shanghai}
  \country{China}
}

\author{Changbo Wang}
\authornotemark[2]
\email{cbwang@cs.ecnu.edu.cn}
\affiliation{%
  \institution{East China Normal University}
  \city{Shanghai}
  \country{China}
}

\author{Guangtao Zhai}
\email{zhaiguangtao@sjtu.edu.cn}
\affiliation{%
  \institution{Shanghai Jiao Tong University}
  \city{Shanghai}
  \country{China}
}

\renewcommand{\shortauthors}{Li et al.}
\renewcommand\footnotetextcopyrightpermission[1]{}
\settopmatter{printacmref=false} 

\begin{abstract}
The oracle bone inscription (OBI) recognition plays a significant role in understanding the history and culture of ancient China. However, the existing OBI datasets suffer from a long-tail distribution problem, leading to biased performance of OBI recognition models across majority and minority classes. With recent advancements in generative models, OBI synthesis-based data augmentation has become a promising avenue to expand the sample size of minority classes. Unfortunately, current OBI datasets lack large-scale structure-aligned image pairs for generative model training. To address these problems, we first present the Oracle-P15K, a structure-aligned OBI dataset for OBI generation and denoising, consisting of 14,542 images infused with domain knowledge from OBI experts. Second, we propose a diffusion model-based pseudo OBI generator, called OBIDiff, to achieve realistic and controllable OBI generation. Given a clean glyph image and a target rubbing-style image, it can effectively transfer the noise style of the original rubbing to the glyph image. Extensive experiments on OBI downstream tasks and user preference studies show the effectiveness of the proposed Oracle-P15K dataset and demonstrate that OBIDiff can accurately preserve inherent glyph structures while transferring authentic rubbing styles effectively.
\end{abstract}


\begin{teaserfigure}
  \includegraphics[width=\textwidth]{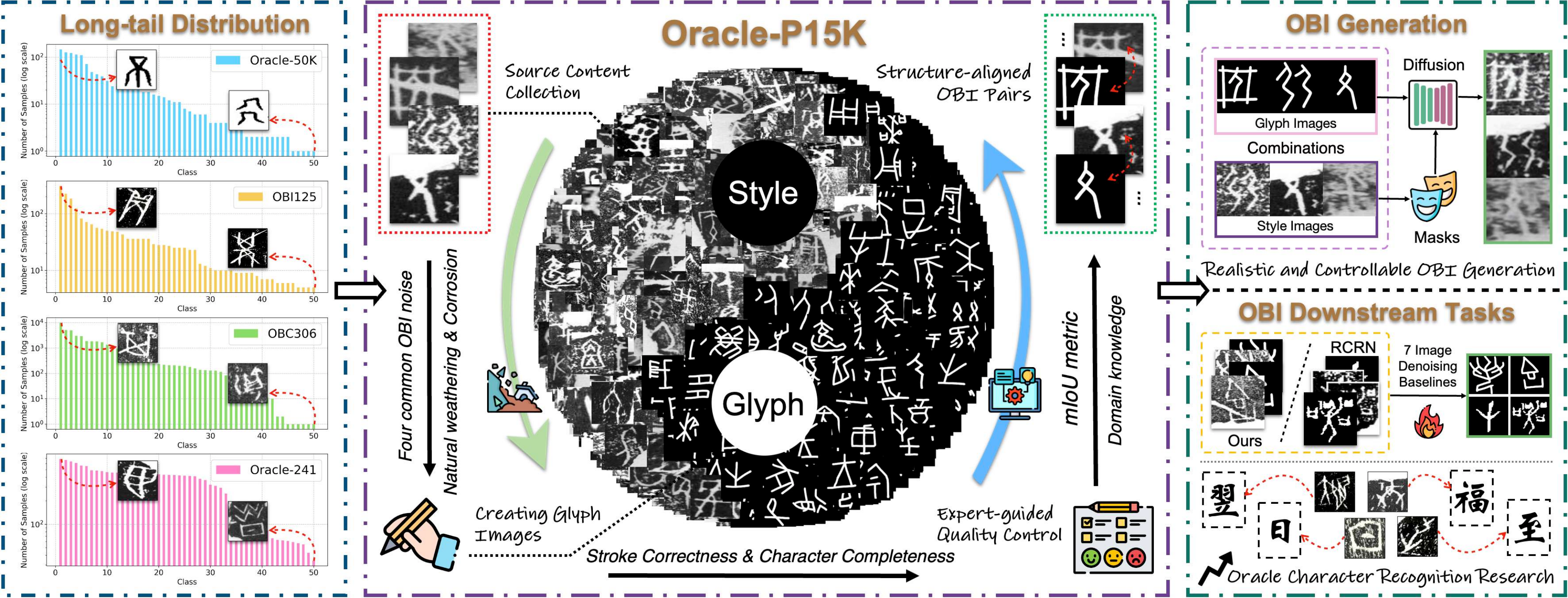}
  \caption{Overview of the proposed Oracle-P15K dataset. The dataset comprises 14,542 OBI images with structure-aligned expert-annotated glyphs. Based on this, we present a pseudo OBI image generator, namely OBIDiff, to alleviate the long-tail distribution problem in current OBI datasets. Extensive experiments demonstrate both the necessity of Oracle-P15K and the effectiveness of OBIDiff in improving the performance of downstream OBI tasks.}
  \label{fig:teaser}
\end{teaserfigure}


\maketitle


\section{Introduction}
\label{sec:intro}

Oracle bone inscriptions (OBI), dating from the late Shang Dynasty (1400–1100 B.C.), are the earliest evidence of Chinese writing. They document divination practices and prayers to deities, offering valuable insights into the language, society, and beliefs of ancient China. The discovery and study of OBI provide a unique perspective on early Chinese civilization. With the advancements of artificial intelligence, various models have been proposed for different OBI information processing tasks, such as OBI recognition \cite{wang2022unsupervised}, denoising \cite{shi2022charformer, shi2022rcrn}, deciphering \cite{guan2024deciphering, chen2024obi, wang2024puzzle, ghaboura2025time, shih2025reasoning, jiang2024oraclesage}, etc. The OBI datasets are essential for these deep learning-based methods. However, most existing OBI datasets suffer from a long-tail distribution problem. As shown in \cref{fig:teaser}, the largest class in the OBC306 \cite{huang2019obc306} contains 51,796 samples, while the smallest class only contains 2 samples. Consequently, OBI-related models achieve superior performance in majority classes while underperforming in minority classes.

To overcome this issue, three strategies have been proposed for the long-tail distribution problem in OBI datasets: few/zero-shot learning, imbalanced learning, and cross-modal learning \cite{li2024comprehensive}. Few/zero-shot learning and imbalanced learning \cite{han2020self, zhao2022ffd, diao2023toward, diao2022rzcr} tackle the challenge of data scarcity with network improvement and data augmentation. However, network improvement is constrained by writing variability and severe noise in OBI. Cross-modal learning \cite{wang2022unsupervised, wang2024oracle} mitigates writing variability by leveraging both scanned and handprint data but still struggles to handle noisy images and address more complex scenarios. Therefore, various researchers use data augmentation \cite{dazheng2021random, li2021mix, wang2022improving, huang2022agtgan, li2023towards, chen2024study} to expand the training set with the most powerful generative models, i.e., generative adversarial networks \cite{goodfellow2014generative, zhu2017unpaired} and diffusion models \cite{ho2020denoising, rombach2022high, zhang2023adding}. For example, AGTGAN \cite{huang2022agtgan} proposed an unsupervised generative adversarial network to generate characters with diverse glyphs and realistic textures, thereby supplementing the dataset with scanned OBI images. ADA \cite{li2023towards} introduced an end-to-end generative adversarial framework that produces synthetic data through convex combinations of all available samples in tail classes. However, these methods suffer from low quality and a lack of controllability, which limits their applications in fine-grained OBI recognition tasks.

In this paper, we aim to develop a method to generate realistic and controllable OBI images by incorporating glyphs and styles. The glyphs indicate the semantics of the OBI and the styles describe the unique texture caused by thousands of years of natural weathering and corrosion. However, training such a model requires large-scale structure-aligned glyph and style image pairs. To the best of our knowledge, none of the existing OBI datasets can provide such labor-intensive and expertise-demanding pairs. The Oracle-241 \cite{wang2022unsupervised} dataset contains OBI image pairs, which are only class-aligned. Shi \emph{et al.} \cite{shi2022rcrn} proposes a structure-aligned dataset for OBI denoising, i.e., RCRN, which is limited in scale and quality. Therefore, we construct Oracle-P15K, a large-scale structure-aligned OBI dataset comprising 14,542 images infused with domain knowledge from OBI experts. The Oracle-P15K dataset can also serve as a comprehensive benchmark for researchers to develop and evaluate their methods for dealing with other OBI information processing tasks, such as OBI denoising, recognition, etc.

Based on our Oracle-P15K, we propose a novel diffusion model for realistic and controllable OBI generation, namely OBIDiff. It can transfer the style of the glyph image into that of the style image. Our model consists of a frozen autoencoder, a pre-trained stable diffusion (SD) model, a trainable glyph encoder, and a trainable style encoder. Unlike natural image styles, OBI image styles are too complicated to define by a few text prompts. Therefore, the style encoder integrates a CLIP \cite{radford2021learning} image encoder and a linear layer to align the image and text embeddings in the latent space. In the training process, the glyph and style encoders extract glyph and style conditions to guide the SD model in generating the target image. It is worth noting that the characters in the glyph and style images should be the same. As a result, the glyph of the style image hampers the training of the glyph encoder. To tackle this problem, we leverage the structure-aligned glyph image to mask the glyph of the style image. In the inference process, the characters in the glyph and style images are different so that we can transfer the style of the glyph image into that of the style image. In conclusion, our key contributions can be summarized as follows:
\begin{itemize}
    \item \textbf{Introduction of Oracle-P15K:} We construct Oracle-P15K, the first OBI dataset containing structure-aligned image pairs, specifically for pseudo OBI generation. This dataset consists of 14,542 images across 239 classes that cover four commonly existing types of noise in OBI.
    \item \textbf{Controllable Pseudo OBI Generator:} We present a dedicated diffusion model, namely OBIDiff, to address the huge sample size discrepancies in current OBI datasets for mitigating the long-tail distribution problem that hinders downstream OBI tasks. By controllably combining learned OBI noise representations into glyph generation, OBIDiff achieves realistic and expert-confused OBI generation.
    \item \textbf{Comprehensive Downstream Task Evaluation:} We demonstrate the contribution of Oracle-P15K to improving the performance of downstream OBI tasks. Specifically, it enhances the capability of OBI denoising methods to restore severely distorted oracle bone characters.
\end{itemize}


\section{Related Work}
\label{sec:related_work}

\subsection{Oracle Bone Inscription Datasets}

\begin{table*}[t]
    \caption{Summary of the existing oracle bone inscription datasets. The Min, Max, and SD denote the minimum, maximum, and standard deviation of sample counts per class in the dataset, respectively. Specifically, we report the distribution information of the test set of our Oracle-P15K separately. The alignment indicates whether the dataset is structure-aligned or not.}
    \label{tab:statistic}
    \begin{tabular}{@{}cccccccccccc@{}}
        \toprule
        Type & Dataset & Year & \#Classes & \#Images & Resolution & Min & Max & SD & Availability & Alignment \\ 
        \midrule
        \multirow{12}{*}{Unpaired} & Oracle-20K \cite{guo2015building} & 2016 & 261 & 20,039 & 50$\times$50 & - & - & - & {\color{red}\ding{55}} & {\color{red}\ding{55}}\\
         & Oracle-50K \cite{han2020self} & 2020 & 2,668 & 59,081 & 50$\times$50 & 1 & 388 & 39.1 & {\color{green}\ding{51}} & {\color{red}\ding{55}}\\
         & HWOBC \cite{li2020hwobc} & 2020 & 3,881 & 83,245 & 400$\times$400 & 19 & 24 & 1.2 & {\color{green}\ding{51}} & {\color{red}\ding{55}}\\
         & OBC306 \cite{huang2019obc306} & 2019 & 306 & 309,551 & $<$382$\times$478 & 1 & 25,898 & 2,428.3 & {\color{green}\ding{51}} & {\color{red}\ding{55}}\\
         & OracleBone8000 \cite{zhang2021ai} & 2020 & - & 129,770 & - & - & - & - & {\color{red}\ding{55}} & {\color{red}\ding{55}}\\
         & OBI125 \cite{yue2022dynamic} & 2022 & 125 & 4,257 & $<$278$\times$473 & 5 & 307 & 42.3 & {\color{green}\ding{51}} & {\color{red}\ding{55}}\\
         & OB-Rejoin \cite{zhang2022data} & 2022 & - & 998 & $<$1408$\times$1049 & - & - & - & {\color{red}\ding{55}} & {\color{red}\ding{55}}\\
         & Oracle-MNIST \cite{wang2024dataset} & 2023 & 10 & 30,222 & 28$\times$28 & 2,628 & 3,699 & 351.1& {\color{green}\ding{51}} & {\color{red}\ding{55}}\\
         & EVOBC \cite{guan2024open} & 2024 & 13,714 & 229,170 & $<$465$\times$857 & 0 & 1,000 & 48.0 & {\color{green}\ding{51}} & {\color{red}\ding{55}}\\
         & HUST-OBC \cite{wang2024open} & 2024 & 10,999 & 140,053 & $<$400$\times$520 & 1 & 307 & 39.6 & {\color{green}\ding{51}} & {\color{red}\ding{55}}\\
         & O2BR \cite{chen2024obi} & 2025 & - & 800 & $<$2664$\times$2167 & - & - & - & {\color{green}\ding{51}} & {\color{red}\ding{55}}\\
         & OBI-rejoin \cite{chen2024obi} & 2025 & - & 200 & $<$2913$\times$1268 & - & - & - & {\color{green}\ding{51}} & {\color{red}\ding{55}}\\
        \midrule
        \multirow{3}{*}{Paired} & Oracle-241 \cite{wang2022unsupervised} & 2022 & 241 & 78,565 & $<$588$\times$700 & 29 & 687 & 175.6 & {\color{green}\ding{51}} & {\color{red}\ding{55}}\\
         & RCRN \cite{shi2022rcrn} & 2022 & - & 3,212 & $<$520$\times$668 & - & - & - & {\color{green}\ding{51}} & {\color{green}\ding{51}}\\
         & Oracle-P15K & 2025 & 239 & 14,542 & 128$\times$128 & 60/22 & 60/114 & 0/23.8 & {\color{green}\ding{51}} & {\color{green}\ding{51}}\\
        \bottomrule
    \end{tabular}
\end{table*}

In the past decade, various oracle bone inscription (OBI) datasets have been established for different OBI information processing tasks. The existing OBI datasets can be categorized into rubbings and handprints from the perspective of content sources. For instance, the OBC306 \cite{huang2019obc306}, OBI125 \cite{yue2022dynamic}, Oracle-MNIST \cite{wang2024dataset} and O2BR \cite{chen2024obi} are rubbing datasets for OBI recognition, whose images are collected from oracle bone publications or shot from real OBI materials. In contrast, the Oracle-20K \cite{guo2015building}, Oracle-50K \cite{han2020self}, and HWOBC \cite{li2020hwobc} are handprint datasets for OBI recognition, whose images are imitatively written by experts. Meanwhile, the Oracle-241 \cite{wang2022unsupervised} is another OBI recognition dataset containing handprints and rubbings. Besides, the OracleBone8000 \cite{zhang2021ai}, OB-Rejoin \cite{zhang2022data}, and OBI-rejoin \cite{chen2024obi} are built for OBI rejoining. Moreover, the EVOBC \cite{guan2024open} and HUST-OBC \cite{wang2024open} are proposed to understand the evolution of OBI. With the advent of diffusion models and large multimodal models (LMMs), they show great potential for OBI deciphering. However, these datasets cannot provide large-scale structure-aligned image pairs, which hinders the development of OBI generation and denoising. A detailed comparison of these datasets is provided in \cref{tab:statistic}.

\subsection{Oracle Bone Inscription Generation}

Traditional oracle bone inscription generation methods augment samples through transformations, such as random rotations and flipping. Recently, RPC \cite{dazheng2021random} generates additional images by mimicking noise, i.e., dense white regions, with various polygons. In addition, MA \cite{li2021mix} leverages feature information from both majority and minority classes to augment samples in minority classes, thereby achieving better recognition performance than traditional methods. Moreover, Orc-Bert \cite{han2020self} and FFD \cite{zhao2022ffd} focus on the few-shot learning of handprint data and augment the few-shot classes by learning and transforming the stroke vectors of OBI, which may be hampered by the severe noise in rubbing data. More recently, CDA \cite{wang2022improving}, AGTGAN \cite{huang2022agtgan}, ADA \cite{li2023towards}, STSN \cite{wang2022unsupervised}, and UDCN \cite{wang2024oracle} consider OBI generation an image translation problem and leverage generative adversarial networks to swap texture features across domains. However, they suffer from a lack of controllability and training instability. In contrast, our OBIDiff can generate realistic OBI images while being controllable.

\subsection{Downstream Oracle Bone Inscription Tasks}

Downstream oracle bone inscription tasks include challenges in six domains: rejoining, classification, retrieval, deciphering, denoising, and recognition \cite{chen2024obi}. Among them, OBI rejoining \cite{zhang2021ai, zhang2022data} restores the original appearance of oracle bones and offers complete and accurate information for OBI researchers. In addition, OBI retrieval \cite{hu2024component, liu2020oracle, ding2024oracle, jiang2023oraclepoints} collects similar OBI samples from existing datasets, thereby facilitating more large-scale database construction. Moreover, OBI classification helps distinguish real rubbings from distortions, which is a practical requirement for human
experts. Most recently, various OBI deciphering models \cite{guan2024deciphering, chen2024obi, wang2024puzzle, ghaboura2025time, shih2025reasoning, jiang2024oraclesage, xie2024diffobi} have also been proposed to understand these ancient Chinese characters by leveraging diffusion models and LMMs. Overall, all these tasks rely on OBI recognition \cite{zhou1995method, meng2017recognition, wang2022unsupervised}, which suffers from the long-tail distribution problem in current OBI datasets. Recent advancements in generative models offer promising avenues for addressing this issue through data augmentation. However, training such models typically requires structure-aligned image pairs. Similarly, OBI denoising \cite{shi2022rcrn, shi2022charformer, jiang2023oraclepoints} also relies on such pairs to effectively remove noise while preserving glyph structures. Hence, building a structure-aligned dataset for both OBI generation and denoising is an urgent need in the OBI research community.


\section{The Oracle-P15K Dataset}

\begin{figure*}[t]
    \includegraphics[width=\linewidth]{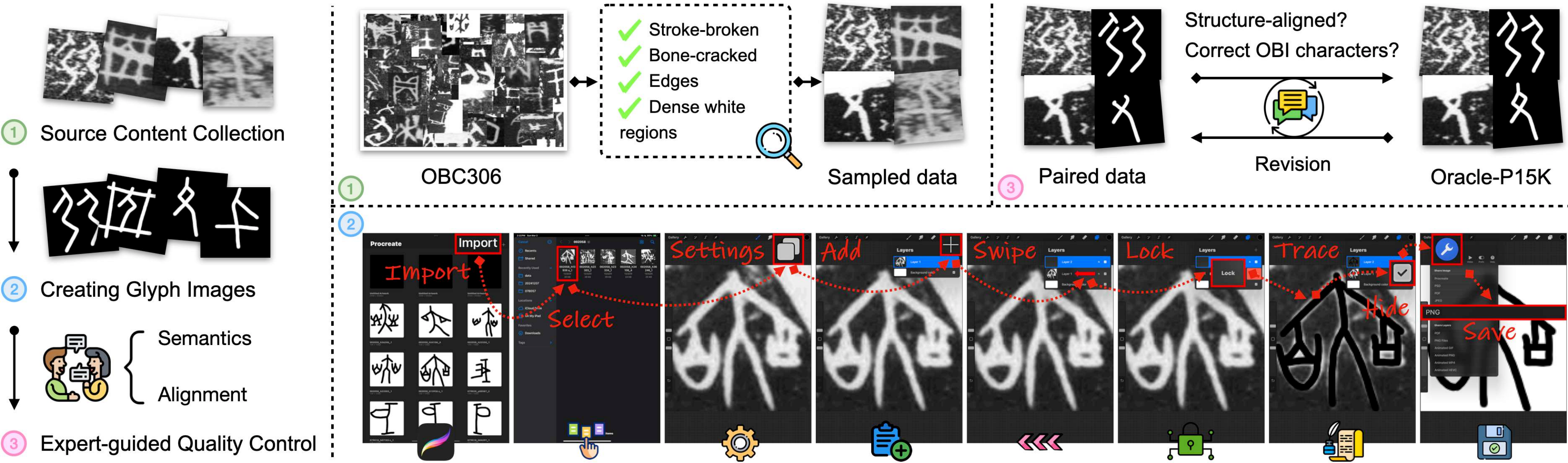}
    \caption{Construction pipeline of our Oracle-P15K. Step-1: We focus on four types of noise in OBI rubbings and take them as targets for source content sampling. Step-2: OBI experts are invited to create glyph images manually. Step-3: We conduct a post-quality examination to ensure the reliability and alignment of OBI image pairs.}
    \label{fig:pipeline}
\end{figure*}

\subsection{Construction Pipeline}
\label{sec:construction}

\cref{fig:pipeline} shows the overall construction pipeline of our Oracle-P15K. The following explains how we build the dataset in detail.

\noindent\textbf{Source Content Collection:} We collect style images from the OBC306 \cite{huang2019obc306}, a large-scale OBI dataset specifically constructed for OBI recognition. According to OBC306, the noise of OBI can be categorized into stroke-broken, bone-cracked, edges, and dense white regions. Thus, we meticulously select representative OBI images from the OBC306 to ensure the coverage of the four noise types in OBI. Specifically, we select 30 images for each of the 200 head classes in OBC306. In addition, all samples of the 39 tail classes are collected for testing to avoid overlapping. Therefore, the whole dataset contains 7,271 image pairs across 239 classes.

\noindent\textbf{Creating glyph images:} After collecting style images, we use an iPad Pro, an Apple Pencil, and the Procreate software to create the corresponding structure-aligned glyph images. All the glyph images are written using a 7-pixel calligraphy monoline-style brush in Procreate. We provide the operational details in \cref{sec:operation}. Specifically, there are two categories of special cases we may encounter in the writing procedure:
\begin{itemize}
    \item Stroke-broken: Some strokes of OBI are broken due to thousands of years of natural weathering and corrosion. In this case, we reconstruct the characters by completing the broken strokes.
    \item Bone-cracked \& Edges: Some strokes of OBI are incomplete or covered by noisy regions caused by long burial and careless excavation. In this case, we complete the glyphs with domain knowledge from the annotator.
\end{itemize}
It should be noted that the glyph image features black strokes on a white background, which is an inverse color configuration compared to the style image. Therefore, we invert the color configuration of the glyph images after construction. 

\noindent\textbf{Expert-guided Quality Control:} We design a quality examination procedure to ensure the quality of the structure-aligned OBI image pairs. Firstly, we conduct a human evaluation involving two OBI experts to assess whether the handprint characters represent correct OBI characters. Moreover, the characters in the glyph images should be structure-aligned with those in the corresponding style images. The experts record the glyph images that fail to meet the quality standards. Any glyph image recorded by one of the experts is discarded. Finally, the annotator is required to revise all discarded images until they faithfully represent the correct OBI characters and achieve satisfied structure alignment.

Considering the limitation of human evaluation in assessing the structure alignment between glyph and style images, we develop an automatic evaluation program to measure the IoU of glyphs between glyph and style images. Any glyph image is discarded if the IoU between it and its corresponding style image is below 0.8. Finally, our Oracle-P15K achieves an mIoU of 0.865.

\subsection{Dataset Statistics}
\label{sec:statistics}

As shown in \cref{tab:statistic}, we compare the proposed Oracle-P15K with other OBI datasets. It can be observed that: \textbf{(1)} Most datasets have reached a 10,000 scale, providing support for the research of machine learning. \textbf{(2)} Unlike other datasets, which only contain handprints, rubbings, or class-aligned image pairs, our Oracle-P15K consists of structure-aligned image pairs infused with domain knowledge from OBI experts. \textbf{(3)} Our dataset is a distribution-balanced dataset, which prevents bias in model training and evaluation. Specifically, it achieves 60, 60, and 0 in terms of minimum, maximum, and standard deviation of sample counts per class in the training and validation sets, respectively. In summary, while the Oracle-P15K and other OBI datasets are all large-scale, our dataset focuses on structure-aligned OBI image pairs, thereby facilitating the development of robust OBI generation and denoising models.

Empirically, we randomly split the samples selected from the 200 head classes of the OBC306 dataset into two subsets for training and validation in an 8:2 ratio. The test set is collected from the 39 tail classes of the OBC306 dataset and does not overlap with the images in the training and validation sets.

\begin{figure*}[t]
    \centering
    \includegraphics[width=\linewidth]{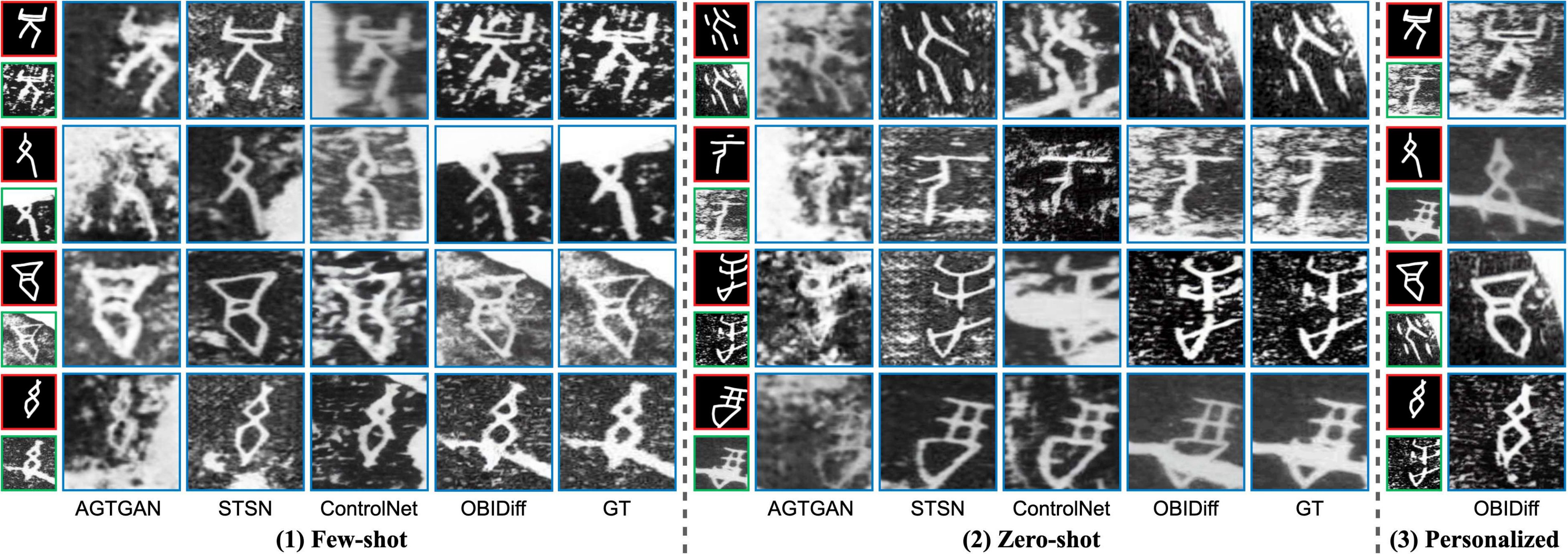}
    \caption{Comparison of the OBI generation results on the Oracle-P15K dataset. The red, green, and blue boxes denote the glyph, style, and generated images, respectively. Three evaluation settings are considered. (1) Few-shot: the characters in the training and validation sets are the same. (2) Zero-shot: the characters in the training and test sets are not overlapped. (3) Personalized: the glyph and style images are not aligned.}
    \label{fig:generation}
\end{figure*}

\begin{figure}[t]
    \centering
    \includegraphics[width=\linewidth]{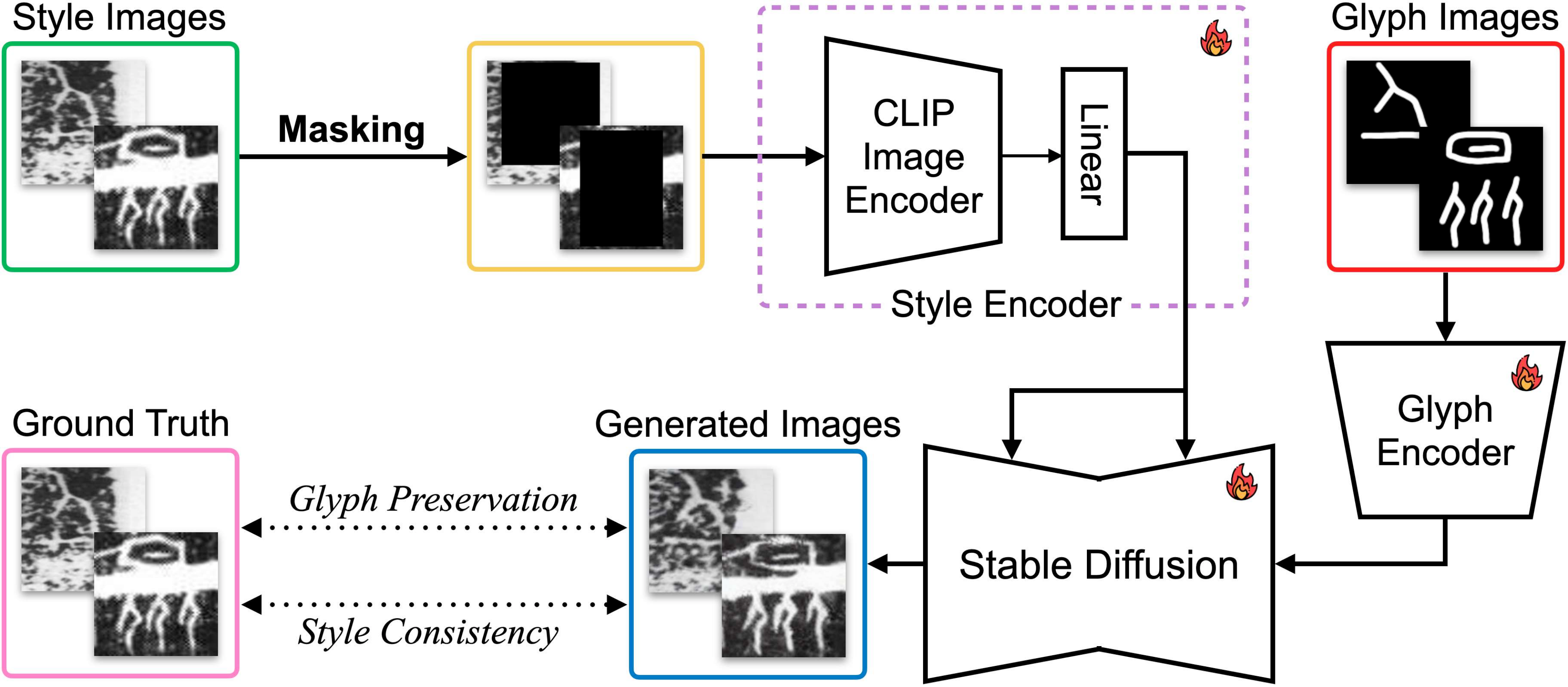}
    \caption{Architecture of our OBIDiff.}
    \label{fig:method}
\end{figure}

\subsection{Application Scenarios}
\label{sec:application}

The Oracle-P15K dataset can contribute to various OBI information processing tasks including, but not limited to, OBI generation and denoising. By enhancing the quality and diversity of OBI images, it improves the robustness and accuracy of OBI recognition models. Specifically, our Oracle-P15K benefits the OBI research community in three folds:
\begin{itemize}
    \item \textbf{Catering OBI Researchers Better:} Existing OBI datasets often contain misannotated images, which can hinder research accuracy. To provide OBI researchers with more fine-grained and reliable data, our Oracle-P15K dataset has undergone multiple rounds of validation by OBI experts.
    \item \textbf{Facilitating OBI Generation Models:} Our dataset enables OBI generation models to generate realistic and controllable OBI images, thereby improving the recognition accuracy of the tail class significantly.
    \item \textbf{More Comprehensive and Practical OBI Denoising Benchmark:} Our dataset contains 7,271 real-world OBI image pairs across 4 types of noise. Compared to the previous OBI denoising benchmark, i.e., RCRN, it is more comprehensive and closer to the real-world OBI denoising scenario.
\end{itemize}

\section{Pseudo OBI Generator}


Inspired by ControlNet \cite{zhang2023adding}, we propose to generate pseudo OBI images by incorporating glyphs and styles. As shown in \cref{fig:method}, our OBIDiff consists of an autoencoder, a stable diffusion (SD) \cite{rombach2022high} model, a glyph encoder, and a style encoder. Firstly, we feed OBIDiff with an input image $x_0$, a glyph image $x_g$, and a style image $x_s$. Then, the autoencoder extracts features from $x_0$ to put the diffusion process into latent space. Subsequently, the glyph conditions $\tau_g$ are extracted by the glyph encoder $\mathcal{E}_g$ to concatenate with initial noise for glyph guidance:
\begin{equation}
    \tau_g = \mathcal{E}_g(x_g).
\end{equation}
In addition to glyph conditions, style conditions play a crucial role in controlling the intricate visual characteristics of the generated images. However, it is difficult to describe the complex styles of OBI images in natural languages. Therefore, we introduce the style encoder $\mathcal{E}_s$ to learn the specific style representations from the real-world OBI rubbing images. The style encoder $\mathcal{E}_s$ comprises a CLIP image encoder $\mathcal{E}_c$ and a linear layer $l$. The CLIP image encoder $\mathcal{E}_c$ extracts image embeddings and the linear layer $l$ aligns them with text embeddings in the latent space:
\begin{equation}
    \tau_s = l(\mathcal{E}_c(x_s)).
\end{equation}
Finally, the SD model, glyph encoder, and style encoder are optimized jointly by the simple loss of latent diffusion models \cite{rombach2022high} computed on the latent noise of the forward process and the reverse process:
\begin{equation}
    \mathcal{L} = \mathbb{E}_{x_0, t, \tau_g, \tau_s, \epsilon\sim\mathcal{N}(0, 1)}\left[\parallel \epsilon - \epsilon_{\theta}(x_t, t, \tau_g, \tau_s)\parallel_2^2\right],
\end{equation}
where $\mathcal{L}$ is the overall learning objective of our OBIDiff. $x_t$ denotes a noisy image produced in time $t$. $\epsilon_t$ represents the learning network, which predicts the noise added to the noisy image $x_t$. Notably, the style image $x_s$ is obtained from the input image $x_0$, which is structure-aligned with the glyph image $x_g$. In the inference process, the style image is obtained from other images. In this case, the style embeddings of the style image guide the generative model in generating a new style OBI image.

\begin{table*}[t]
    \caption{Quantitative comparisons of OBI generation results on the Oracle-P15K dataset. The improvements of our OBIDiff against other baselines are delimited by slashes. The best and second-best results are in {\color{red} red} and {\color{blue} blue}, respectively.}
    \label{tab:quantitative}
    \fontsize{8pt}{11pt}\selectfont
    \begin{tabular}{@{}cccccccccc@{}}
        \toprule
        \multirow{2}{*}{\textbf{Methods}} & \multicolumn{6}{c}{\textbf{Few-shot}} & \multicolumn{3}{c}{\textbf{Zero-shot}} \\
        \cmidrule(r){2-7} \cmidrule(r){8-10} & L1 Loss$\downarrow$ & RMSE$\downarrow$ & PSNR$\uparrow$ & SSIM$\uparrow$ & LPIPS$\downarrow$ & FID$\downarrow$ & Acc@1$\uparrow$ & Acc@3$\uparrow$ & Acc@5$\uparrow$ \\
        \midrule
        Rubbings (ref) & - & - & - & - & - & - & 86.32 & 94.81 & 97.64 \\
        Handprints & 0.3874/0.2682 & 0.4895/0.2881 & 6.2084/7.7200 & 0.1211/0.3628 & 0.5440/0.3591 & 344.1/271.3 & 14.99/61.74 & 28.19/60.85 & 36.69/56.82 \\
        \midrule
        AGTGAN \cite{huang2022agtgan} & 0.3184/0.1992 & 0.3981/0.1967 & 8.0039/5.9245 & 0.0778/0.4061 & 0.5012/0.3163 & 196.0/123.2 & 54.01/24.76 & 71.93/19.11 & 80.42/13.92 \\
        STSN \cite{wang2022unsupervised} & {\color{blue} 0.2648}/0.1456 & {\color{blue} 0.3581}/0.1567 & {\color{blue} 8.9290}/4.9994 & {\color{blue} 0.2486}/0.2353 & {\color{blue} 0.3662}/0.1813 & 110.3/37.5 & {\color{blue} 71.23}/7.54 & 84.20/6.84 & 88.44/5.90 \\
        ControlNet \cite{zhang2023adding} & 0.2877/0.1685 & 0.3743/0.1729 & 8.5403/5.3881 & 0.1776/0.3063 & 0.4147/0.2298 & {\color{blue} 100.0}/27.2 & 70.05/8.72 & {\color{blue} 85.38}/5.66 & {\color{blue} 91.27}/3.07 \\
        \midrule
        OBIDiff & {\color{red} 0.1192} & {\color{red} 0.2014} & {\color{red} 13.9284} & {\color{red} 0.4839} & {\color{red} 0.1849} & {\color{red} 72.8} & {\color{red} 78.77} & {\color{red} 91.04} & {\color{red} 94.34} \\
        \bottomrule
    \end{tabular}
\end{table*}

\begin{figure*}[h]
    \includegraphics[width=\linewidth]{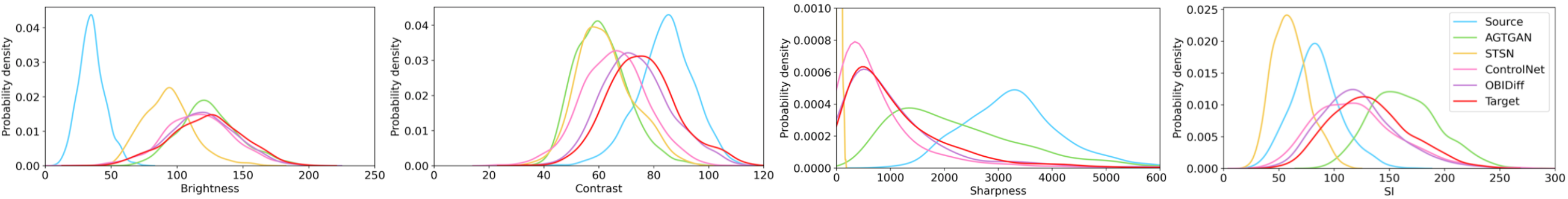}
    \caption{Feature distribution comparisons among generated images from baseline methods and our OBIDiff.}
    \label{fig:features}
\end{figure*}

\begin{figure}[t]
    \includegraphics[width=\linewidth]{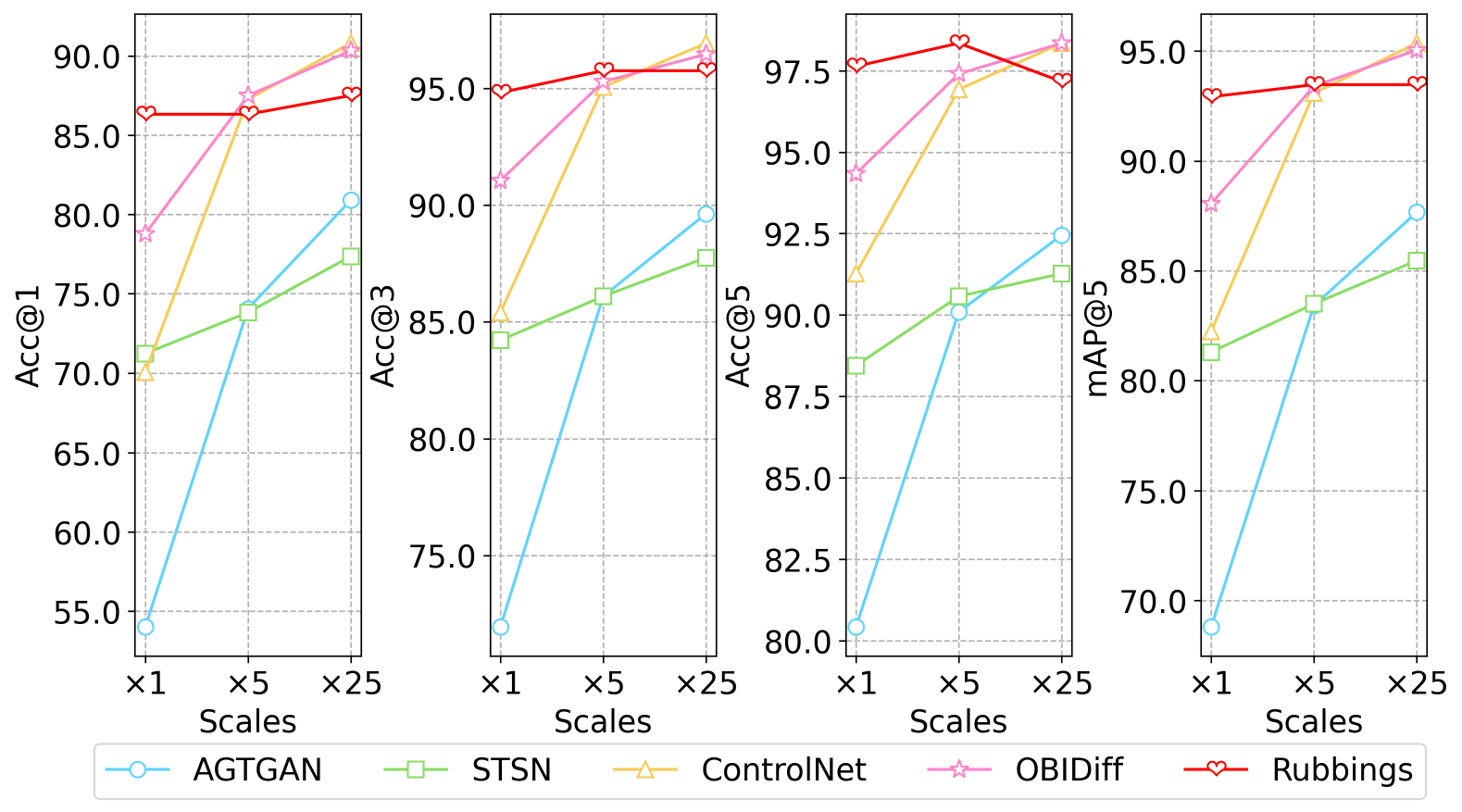}
    \caption{Recognition accuracy of augmented rare OBI images w.r.t. style image at different scales.}
    \label{fig:scales}    
\end{figure}

However, the glyph encoder fails to learn the glyph conditions in the training process. Consequently, the generated images are identical to the style images in the inference process, which results from the glyph interference in the training process. To address this issue, we mask the glyph of the style image with a bounding box computed by detecting the white pixels in the structure-aligned glyph image. This strategy alleviates the influence of the glyph while preserving the main information of the style image, which effectively enhances the capability of the model to learn the glyph and style conditions.


\section{Experiments}

\subsection{Experimental Settings} 

\noindent
{\bf Datasets.} Our OBIDiff is evaluated on the validation and testing sets of the proposed Oracle-P15K dataset. Three generation strategies are considered. First, the characters of the training and validation sets are the same. Second, a zero-shot generation case is formed by performing inference on 39 tail classes from the testing set. Third, we introduce personalized OBI generation by substituting the style images with different characters. Besides,  we collect the other 424 images across 39 tail classes from the test set of the OBC306 to evaluate the models on the OBI recognition task.

To the best of our knowledge, RCRN \cite{shi2022rcrn} is the only existing dataset consisting of structure-aligned OBI image pairs. Therefore, we also conduct parallel experiments on RCRN to demonstrate the superiority of our Oracle-P15K on the OBI denoising task. We collect 4,499 images across 200 head classes from the OBC306 dataset to evaluate their denoising performance. Note that these images and the proposed Oracle-P15K dataset do not overlap.

\noindent
{\bf Baseline Methods.} For the OBI generation task, we compare our OBIDiff with ControlNet \cite{zhang2023adding} and two OBI-specialized image generation models, AGTGAN \cite{huang2022agtgan} and STSN \cite{wang2022unsupervised}. For the OBI denoising task, we benchmark six representative generic image denoising models (DnCNN \cite{zhang2017beyond}, InvDN \cite{liu2021invertible}, Uformer \cite{wang2022uformer}, Restormer \cite{zamir2022restormer}, KBNet \cite{zhang2023kbnet}, and CGNet \cite{ghasemabadi2024cascadedgaze}) as well as an OBI denoising method, CharFormer \cite{shi2022charformer}. All baselines are retrained on OBI task-related datasets using official implementations with default configurations.

\noindent
{\bf Evaluation Metrics.} We apply six common metrics in image generation tasks, including L1 Loss, RMSE, PSNR, SSIM \cite{hore2010image}, LPIPS \cite{zhang2018unreasonable}, FID \cite{heusel2017gans}. 
Since the significant stroke disparity between the handprints and rubbings in our Oracle-P15K, it is irrational to evaluate the OBI denoising performance by standard low-level vision metrics, such as PSNR and SSIM \cite{hore2010image}.
Therefore, an efficient OBI classifier based on ResNet-50 is built to evaluate their performance in tasks oriented to oracle character recognition. More implementation details are in Appendix \ref{sec:implementation}.

\begin{figure*}[ht]
    \includegraphics[width=\linewidth]{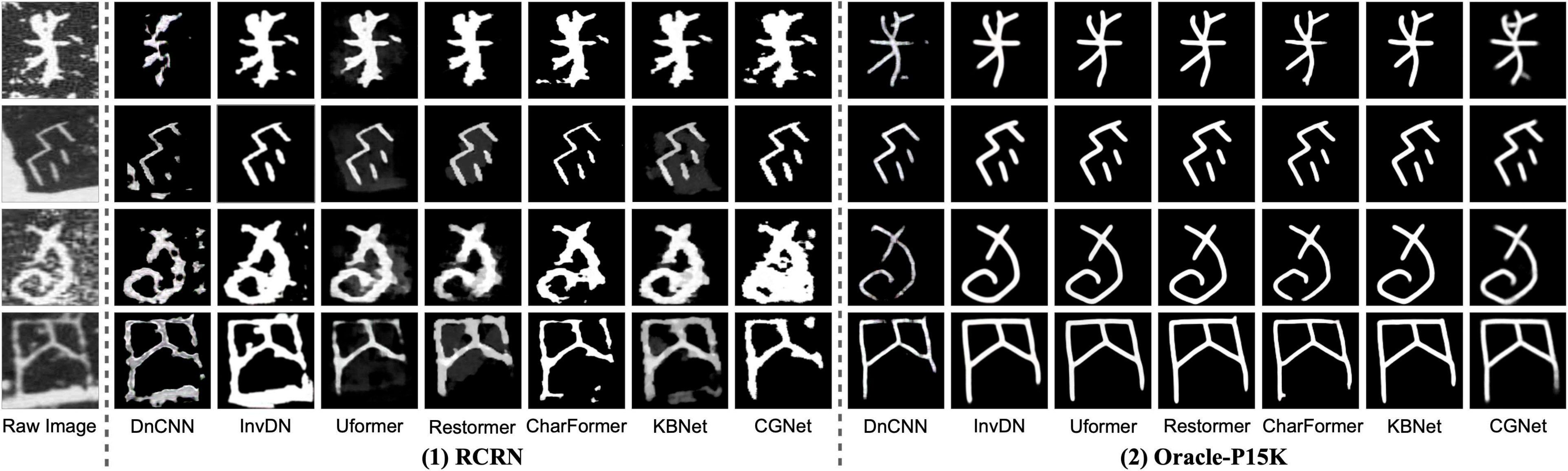}
    \caption{Performance comparison of OBI denoising methods on the RCRN and Oracle-P15K datasets.}
    \label{fig:denoising}
\end{figure*}

\begin{table*}[t]
    \caption{Quantitative comparisons (oracle character recognition accuracy) of OBI denoising methods trained on the RCRN/Oracle-P15K datasets. We also report the improvements of the models trained on our Oracle-P15K against the original rubbings.}
    \label{tab:denoising}
    \fontsize{8pt}{11pt}\selectfont
    \begin{tabular}{@{}ccccccccc@{}}
	\toprule
	  Metrics & Rubbings & DnCNN \cite{zhang2017beyond} & InvDN \cite{liu2021invertible} & Uformer \cite{wang2022uformer} & Restormer \cite{zamir2022restormer} & CharFormer \cite{shi2022charformer} & KBNet \cite{zhang2023kbnet} & CGNet \cite{ghasemabadi2024cascadedgaze} \\ 
        \midrule
        Acc@1$\uparrow$ & 85.97 & 62.06/85.83\textsubscript{-0.14} & 77.28/89.62\textsubscript{+3.65} & 84.71/89.83\textsubscript{+3.86} & 81.42/89.90\textsubscript{+3.93} & 78.19/88.43\textsubscript{+2.46} & 83.17/90.25\textsubscript{+4.28} & 74.05/90.67\textsubscript{+4.70} \\
        Acc@3$\uparrow$ & 95.30 & 76.37/94.18\textsubscript{-1.12} & 88.43/96.70\textsubscript{+1.40} & 94.46/96.56\textsubscript{+1.26} & 92.43/96.84\textsubscript{+1.54} & 89.41/96.21\textsubscript{+0.91} & 93.34/97.05\textsubscript{+1.75} & 86.75/97.62\textsubscript{+2.32} \\
        Acc@5$\uparrow$ & 97.12 & 81.77/96.56\textsubscript{-0.56} & 91.73/98.04\textsubscript{+0.92} & 96.56/98.11\textsubscript{+0.99} & 95.30/98.25\textsubscript{+1.13} & 92.43/97.90\textsubscript{+0.78} & 95.72/98.11\textsubscript{+0.99} & 89.90/98.53\textsubscript{+1.41} \\
        \bottomrule
    \end{tabular}
\end{table*}


\subsection{Main Results}

\noindent
{\bf OBI generation.} \cref{fig:generation} presents the qualitative comparisons between our OBIDiff and other OBI generation methods. We observe that our OBIDiff accurately preserves glyph structures while transferring complex styles effectively. In contrast, other methods struggle to maintain the intrinsic glyphs or reproduce the styles properly. This is primarily because other methods lack effective style controllability. For example, our OBIDiff shows robust generation ability across four different noise types. However, STSN produces duplicated patterns in the third zero-shot case. Moreover, even in the few-shot scenarios, AGTGAN and ControlNet generate artifacted noise in the second and third cases, respectively. More generated results from our OBIDiff can be found in the Appendix.

Additionally, we report the quantitative comparisons between our OBIDiff and other OBI generation methods in \cref{tab:quantitative}. The improvements of our OBIDiff against other OBI generation methods are delimited by slashes. We compare three groups of images: handprint (glyph) images, rubbing (style) images, and generated images. For each glyph image, we generate the augmented image using the corresponding style image. It can be observed that our OBIDiff outperforms other OBI generation methods across all six low-level metrics by a large margin. This can be attributed to the glyph and style encoders, which inject glyph and style information into the SD model to generate realistic and controllable OBI images. Moreover, our OBIDiff achieves state-of-the-art performance in the OBI recognition task. On the one hand, the low recognition accuracy of handprints underscores the significant disparity between handprints and rubbings. We also provide visualization studies in Appendix \ref{sec:vis} to support this conclusion. On the other hand, while ControlNet shows promising results on Acc@3 (85.38\%) and Acc@5 (91.27\%), it performs poorly in terms of Acc@1 (70.05\%). In comparison, our OBIDiff can improve Acc@1 by 8.72\% and 7.54\% against ControlNet and STSN, respectively. These results demonstrate the substantial impact of our OBIDiff in enhancing the recognition accuracy for zero-shot classes. Furthermore, we calculate four low-level features including brightness, contrast, sharpness, and spatial information (SI), thereby providing a large visual space in which to plot and analyze content diversities of the generated images from different models. \cref{fig:features} shows the fitted kernel distribution of each selected feature. Compared to other images, the images generated by our OBIDiff exhibit a more similar probability density. It indicates that our OBIDiff generates more visually pleasant results.

\begin{figure}[t]
    \includegraphics[width=\linewidth]{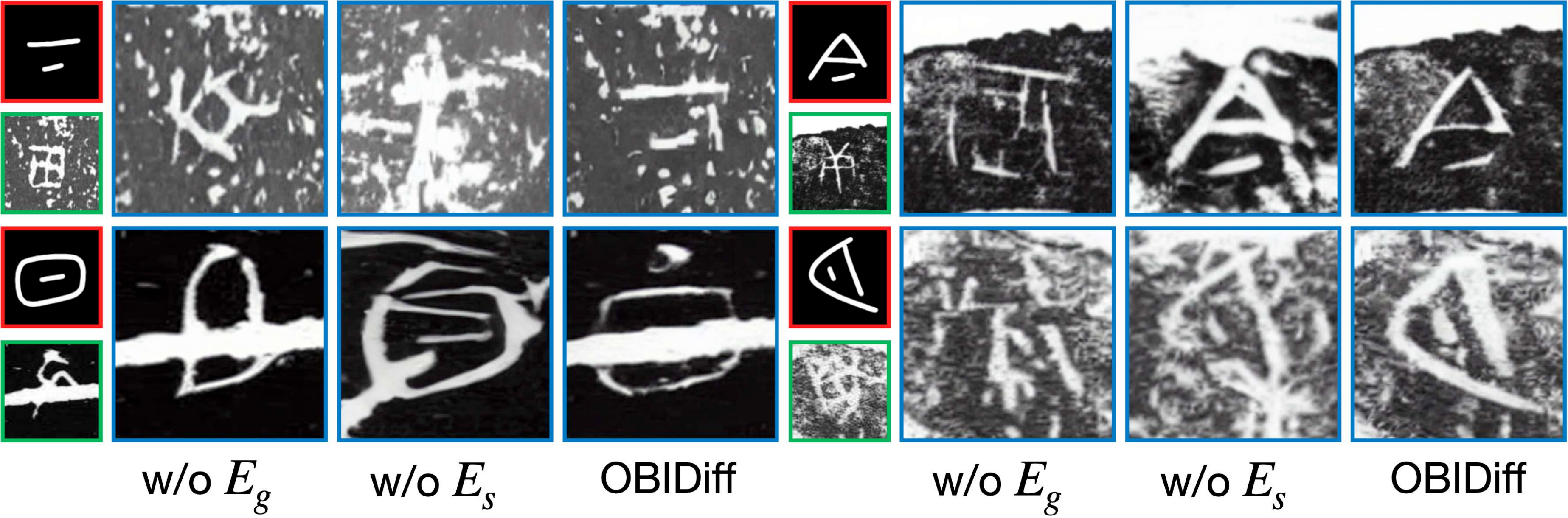}
    \caption{The ablation study of glyph and style encoders.}
    \label{fig:guidance}
\end{figure}

Besides, we investigate the impact of data augmentation by adjusting the scales of data augmentation. \cref{fig:scales} presents the recognition accuracy of augmented images generated by the proposed OBIDiff and other OBI generation methods w.r.t. the scale of data augmentation. All models show consistent progress as the scale of data augmentation increases, except for the original rubbings, which experience a notable decline in Acc@5 at $\times 25$ scale. That may be due to overfitting caused by the duplicated images. Moreover, we also find that ControlNet and our OBIDiff exhibit remarkable improvements as the scale of data augmentation increases. Compared to ControlNet, our OBIDiff achieves superior performance at $\times 1$ scale, narrowing the gap between ControlNet and the original rubbings by half. Trained on our Oracle-P15K, ControlNet surpasses original rubbings by 1.18\% in terms of Acc@3 and Acc@5 at $\times 25$ scale, underscoring the superior quality of the proposed Oracle-P15K. Similarly, our OBIDiff yields a favorable gain of 0.71\% and 1.18\% over the original rubbings, which demonstrates the high quality and diversity of the generated images from our OBIDiff. Overall, our OBIDiff achieves state-of-the-art performance in mAP@5, outperforming the original rubbings by 1.58\% at $\times 25$ scale.

\noindent{\bf OBI denoising.} \cref{fig:denoising} shows qualitative results of generic and OBI denoising models on RCRN and Oracle-P15K datasets. We report their denoising performance on four different noise types. It can be observed that the models trained on Oracle-P15K can robustly handle diverse noise types, whereas the models trained on RCRN struggle to preserve inherent glyph structures properly. For instance, DnCNN and InvDN fail to effectively restore the dense white regions and bone-cracked. Moreover, other methods remove the noise aggressively, leading to glyph damage in the fourth case of \cref{fig:denoising}. Overall, it is demonstrated that the proposed Oracle-P15K dataset can support real-world OBI denoising with large-scale structure-aligned image pairs. More denoising results from different baselines can be found in the Appendix.

In addition, we evaluate the quantitative performance of these methods. As shown in \cref{tab:denoising}, CGNet achieves the best recognition accuracy (90.67\%, 97.62\%, and 98.53\% in terms of Acc@1, Acc@3, and Acc@5, respectively) on Oracle-P15K and Uformer achieves the best recognition accuracy (84.71\%, 94.46\%, and 96.56\% in terms of Acc@1, Acc@3, and Acc@5, respectively) on RCRN. It is an interesting finding that the recognition accuracy of images denoised by the models trained on RCRN is lower than that of the original rubbings. That is due to the significant disparity in image styles between RCRN and the real-world OBI rubbing images. Moreover, we can also find that DnCNN trained on Oracle-P15K achieves poor performance on the OBI recognition task while the other methods enhance the recognition accuracy. Notably, CGNet yields 4.70\%, 2.32\%, and 1.41\% improvements against the original rubbings in terms of Acc@1, Acc@3, and Acc@5, respectively. It highlights the superior quality of our Oracle-P15K and the importance of OBI denoising in OBI recognition.

\subsection{Ablation Studies}

\noindent\textbf{Validation of glyph guidance and style guidance:} We study the function of glyph guidance and style guidance by freezing the parameters of the glyph encoder $\mathcal{E}_g$ and the style encoder $\mathcal{E}_s$ in the training process, respectively. The ablation results are reported in \cref{fig:guidance}. It can be observed that the model generates meaningless glyph structures without the optimization of the glyph encoder. Also, the styles of generated results are uncontrollable if we freeze the parameters of the style encoder. In contrast, our OBIDiff preserves inherent glyphs while transferring the given styles effectively. Overall, our glyph and style encoders contribute to significant gains in quality and controllability.

\begin{figure}[t]
    \includegraphics[width=\linewidth]{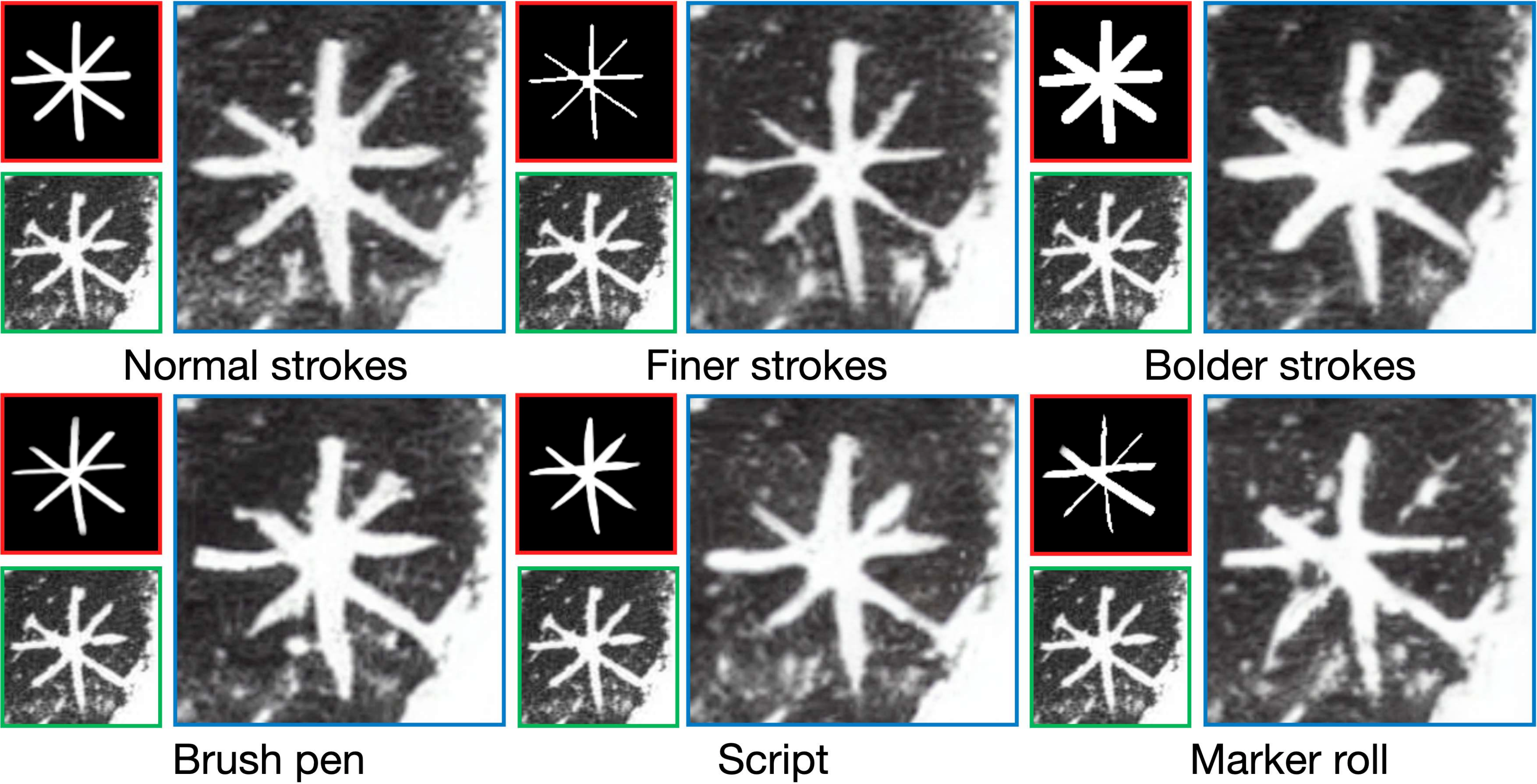}
    \caption{The ablation study of different glyph variations.}
    \label{fig:variations}
\end{figure}

In addition, we also explore the impact of different variations in glyph images. Specifically, we incorporate glyph images with different stroke thicknesses and writing styles in the inference process. As shown in \cref{fig:variations}, our OBIDiff generates images robustly across diverse glyph image variations, thereby ensuring its applicability to the existing handprint OBI datasets.

\begin{figure}[t]
    \includegraphics[width=\linewidth]{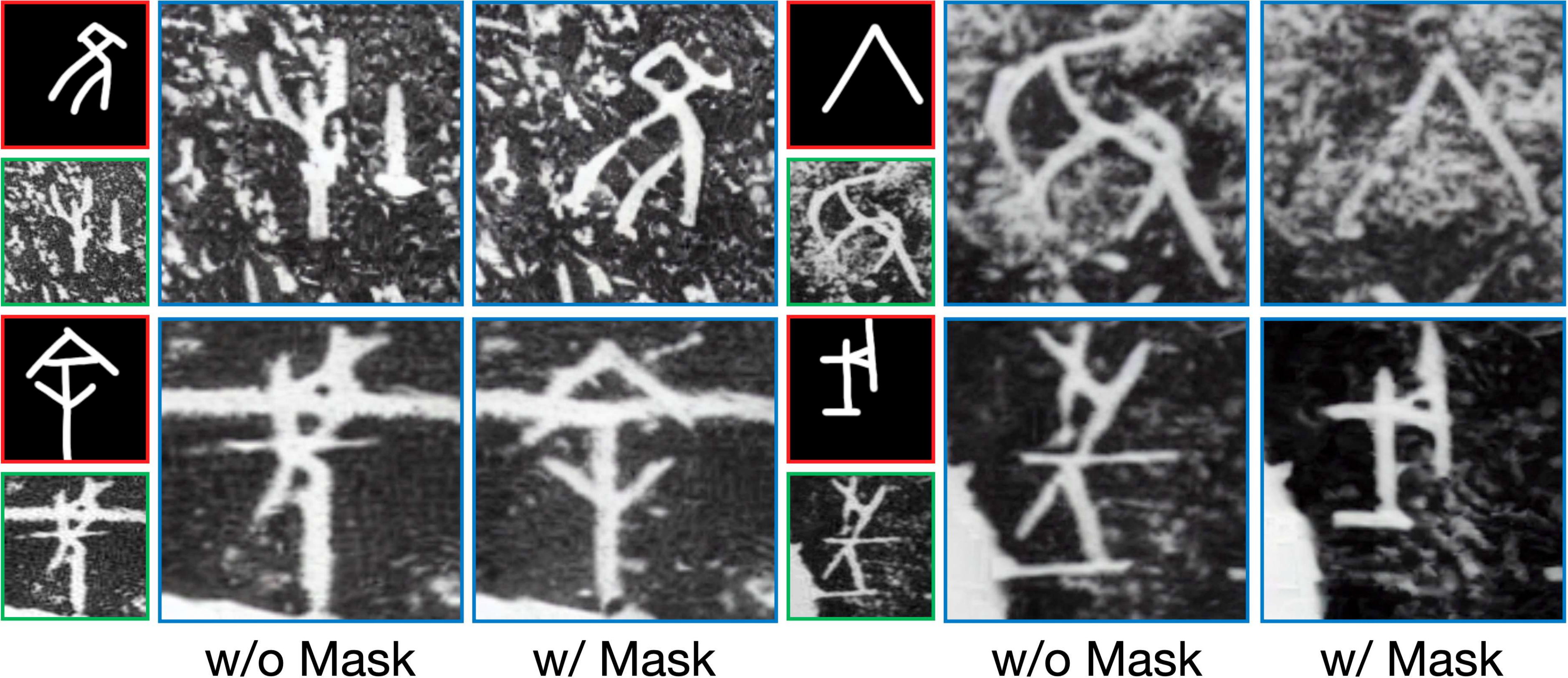}
    \caption{The ablation study of our masking mechanism.}
    \label{fig:mechanism}
\end{figure}

\noindent\textbf{Impact of masking mechanism:} The masking mechanism is designed to mitigate the influence of the glyphs in the style images. We perform an ablation study for the masking mechanism by directly feeding style images to the style encoder. The results are presented in \cref{fig:mechanism}. We observe that, without the masking mechanism, the generated images fail to follow the glyph conditions in the inference process. One possible reason is that the glyphs of the style images hamper the optimization of the glyph encoder in the training process. As a result, the model generates the images by simply copying the style images. Nevertheless, when the model is trained with the masking mechanism, it effectively learns the glyph and style conditions in the training process. Therefore, though the glyphs of the glyph and style images are different in the inference process, it succeeds in transferring the styles while preserving the inherent glyphs robustly.

\subsection{User Preference Study}
We conduct a user preference study to further evaluate the quality of pseudo OBI images generated by our OBIDiff. The study involves 15 college students, most majoring in Chinese language and literature. To facilitate the evaluation process, we develop a web-based user interface with automated navigation. In each round, the interface displays either a ground truth image or a pseudo image generated by our OBIDiff. Participants are asked to determine whether the current image is a real-world OBI rubbing image or a generated image. After completing 100 rounds, they will receive a brief report summarizing their evaluation results. We show the average precision, recall, and F1 score from the user preference study in \cref{tab:user}. It can be observed that the performance in all three metrics is marginally above 0.5 (\textit{random guess}), demonstrating the remarkable similarity between the generated images and real-world rubbing images. The details of the user preference study are shown in Appendix \ref{sec:ui}.
\begin{table}[t]
    \caption{Results of user preference study about comparing the real rubbings and the images generated by OBIDiff.}
    \label{tab:user}
    \begin{tabular}{@{}cccc@{}}
        \toprule
         & Precision$\uparrow$ & Recall$\uparrow$ & F1 score$\uparrow$ \\
        \midrule
       Human Performance & 0.52 & 0.57 & 0.53 \\
        \bottomrule
    \end{tabular}
\end{table}

\subsection{Limitations}
\begin{figure}[t]
    \centering
    \includegraphics[width=\linewidth]{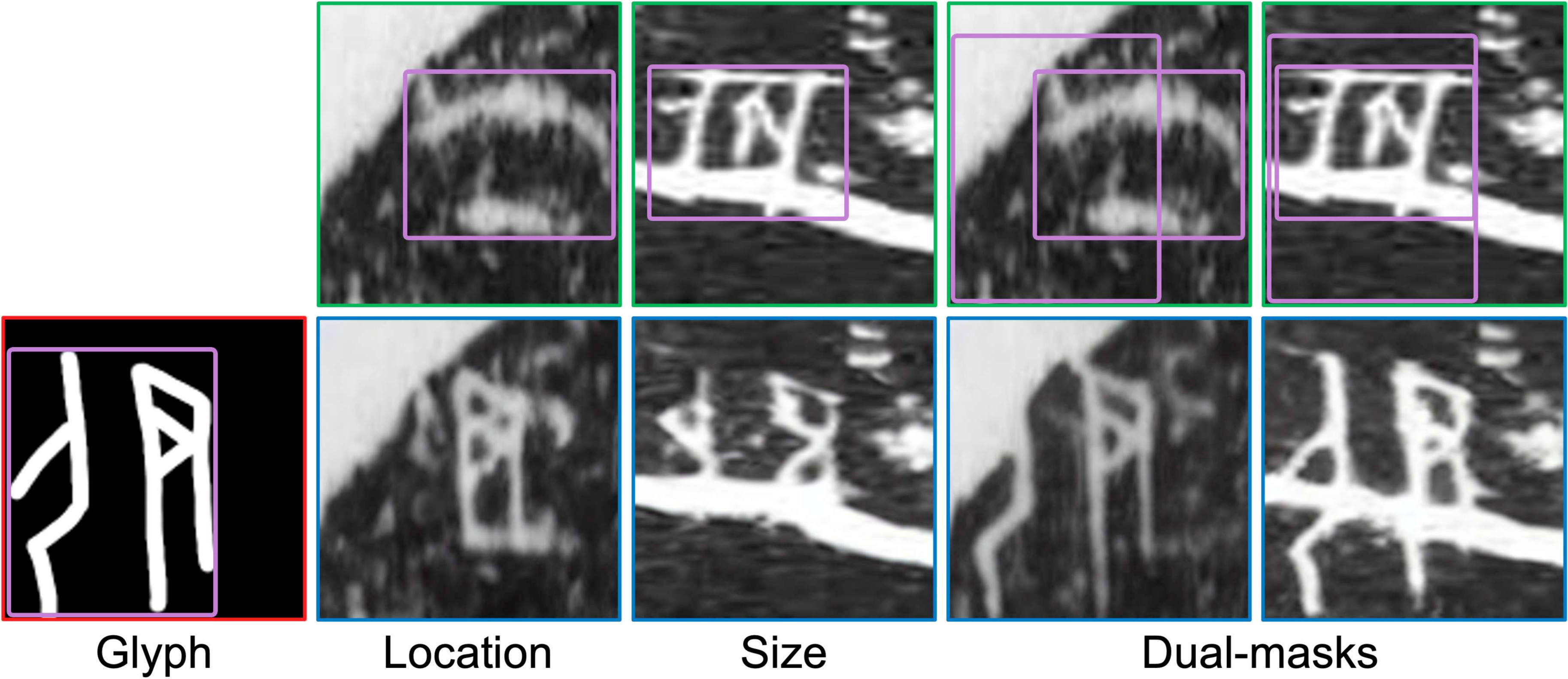}
    \caption{Inconsistency in bounding boxes of glyph and style images. Purple boxes indicate the bounding boxes of the characters. We adopt a dual-masking mechanism to address this issue.}
    \label{fig:limitations}
\end{figure}

Although our method can transfer typical styles to some glyphs, it has two main limitations. First, our model requires a handprint glyph image as one of the inputs, which involves some manual effort. While various handprint OBI datasets are available, this issue still limits its applications in large-scale personalized OBI generation. Second, we introduce a masking mechanism in the style conditioning process to alleviate the influence caused by the glyphs of the style images. However, the location and size of the bounding boxes of the characters in the style images during the inference process may not always be consistent with those in the glyph images. As shown in \cref{fig:limitations}, we adopt a dual-masking mechanism to address this issue. Nevertheless, the mask computed from the glyph image may obscure critical style information in the corresponding style image, leading to a few failed cases in the inference process. Fortunately, this issue can be mitigated by manually selecting the bounding boxes-aligned style images.


\section{Conclusion}

In this paper, we introduce Oracle-P15K, a structure-aligned OBI dataset comprising unique glyph images provided by OBI experts for OBI generation and denoising models. Based on this, we propose a pseudo OBI generator to address the long-tail distribution problem in current OBI datasets, thus significantly improving the recognition accuracy of small-sample characters. Our dataset enables new research directions, such as developing more powerful generative models for rare OBI protection and addressing authenticity concerns like misinformation by identifying fake OBI. Extensive experiments and user preference studies demonstrate the effectiveness of our model on high-quality and noise-controllable OBI generation. Besides, our dataset can also serve as a comprehensive and practical OBI denoising benchmark.

\begin{acks}
This work was supported by the National Social Science Foundation of China (24Z300404220) and the Shanghai Philosophy and Social Science Planning Project (2023BYY003).
\end{acks}

\balance
\clearpage



\balance
\clearpage

\appendix

\section{Ethical Discussions}
\label{sec:ethical}

The Oracle-P15K dataset is crucial for cultural preservation and academic research in the OBI research community. It can contribute to various OBI information processing tasks including, but not limited to, OBI generation and denoising. OBI generation not only facilitates the conservation and study of this invaluable cultural heritage but also enriches academic research with comprehensive material resources. Meanwhile, OBI denoising removes noise to enhance clarity for accurate interpretation and analysis. The integration of these two approaches establishes an innovative paradigm for artificial intelligence applications in cultural heritage restoration, thereby bridging the convergence of traditional civilization and modern technological innovation.

Besides, our dataset might also hold negative social impacts. Firstly, the use of unauthorized or unverified data in the generation process may raise ethical concerns regarding intellectual property rights and ownership claims over cultural heritage. Secondly, over-reliance on algorithmic restoration risks historical misinterpretation or fabrication, potentially undermining academic rigor. Additionally, misusing the technology for artifact forgery or commercial exploitation could compromise cultural authenticity and public trust. Therefore, ethical frameworks must prioritize legal data sourcing, transparent restoration protocols, and respect for the integrity of cultural heritage and the rights of its custodians.

\section{Operational Details}
\label{sec:operation}

Firstly, we import the style image by clicking the import button in the main interface of Procreate. The pop-up displays the directory of the style images, where we select the target image. Subsequently, we click the settings button to show the information of different layers. We need to add a new layer so that the drafts we write can be saved separately. Then, we swipe the attribute bar of the original layer to the left and lock it by clicking the lock button on the right. After that, the glyph is meticulously traced over the style image. Once the writing process is complete, the original layer is hidden, and the newly added layer is saved as the final glyph image.

\section{Implementation Details}
\label{sec:implementation}

\subsection{Implementation Details of OBIDiff}
\label{sec:obidiff}

The proposed OBIDiff is trained on an NVIDIA RTX 4090 GPU for 200 epochs using the AdamW optimizer with a weight decay of 0.01, $\beta_1=0.9$, and $\beta_2=0.99$. During training, the learning rate is set to 1e-5. We use a batch size of 1, and the size of each image is set to 128 $\times$ 128. The entire training process spans over 2 days.

\subsection{Implementation Details of Classifier}
\label{sec:classifier}

Our custom-built classifier is built upon ResNet-50. The model is implemented using the PyTorch platform and trained on an NVIDIA RTX 4090 GPU for 500 epochs with the Adam optimizer. The learning rate is set to 5e-5 and the weight decay is set to 0.01. We use a batch size of 512, and the size of each image is set to 128 $\times$ 128. Random rotation and horizontal flipping techniques are used for data augmentation.

\section{Visualizations}
\label{sec:vis}

\begin{figure*}[ht]
    \centering
    \includegraphics[width=\linewidth]{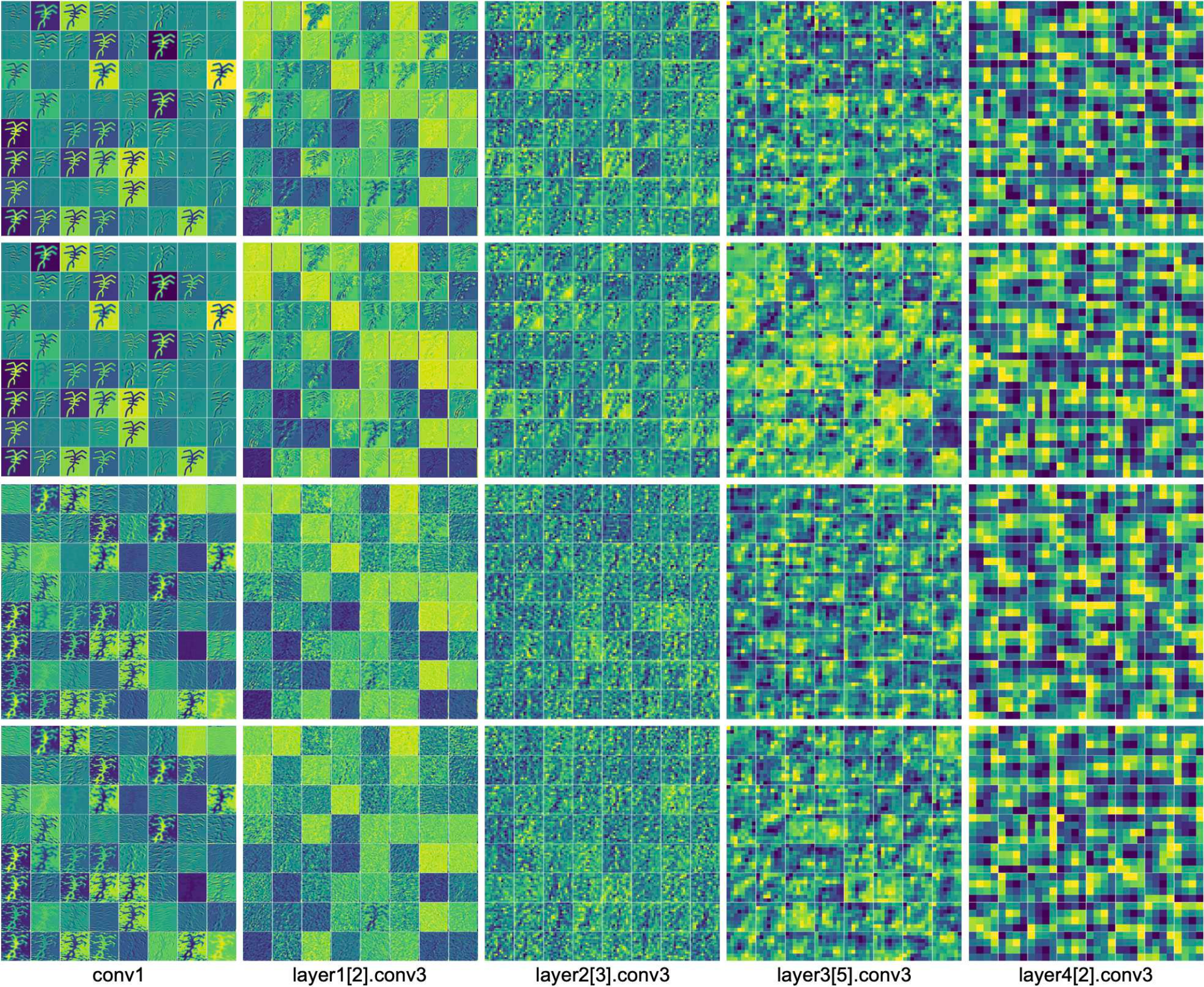}
    \caption{Feature map visualizations of ResNet-50. From top to bottom we present the visualizations of feature maps extracted from the handprint and rubbing images by the models trained on handprint and rubbing OBI datasets, respectively. For each image, we visualize the feature map of each channel.}
    \label{fig:vis}
\end{figure*}

We conduct feature map visualizations using \textit{matplotlib} to analyze the hierarchical feature extraction patterns in convolutional neural networks. As shown in \cref{fig:vis}, we systematically compare the activation patterns of both the initial and final convolutional layers within each residual stage of ResNet-50. Through comparative visualizations of feature maps generated by separately trained models on handprint and rubbing OBI datasets, we have two key observations. First, shallow-layer features (e.g., conv1 and final conv layer of stage 1) exhibit \textit{striking structural similarities} between different models, particularly in edge detection and basic texture response patterns. Second, deeper layers (final conv layers of stage 3-4) demonstrate \textit{divergent specialization}. The handprints-trained model develops enhanced sensitivity to stroke endings and bifurcations, while the rubbings-trained model prioritizes texture continuity. This progressive divergence suggests that while early convolutional layers learn domain-agnostic visual primitives, deeper layers evolve task-specific representations through residual learning. Therefore, the disparity in handprint and rubbing OBI images has a substantial impact on OBI recognition.

\section{Details of the User Preference Study}
\label{sec:ui}

\begin{figure}[H]
    \includegraphics[width=\linewidth]{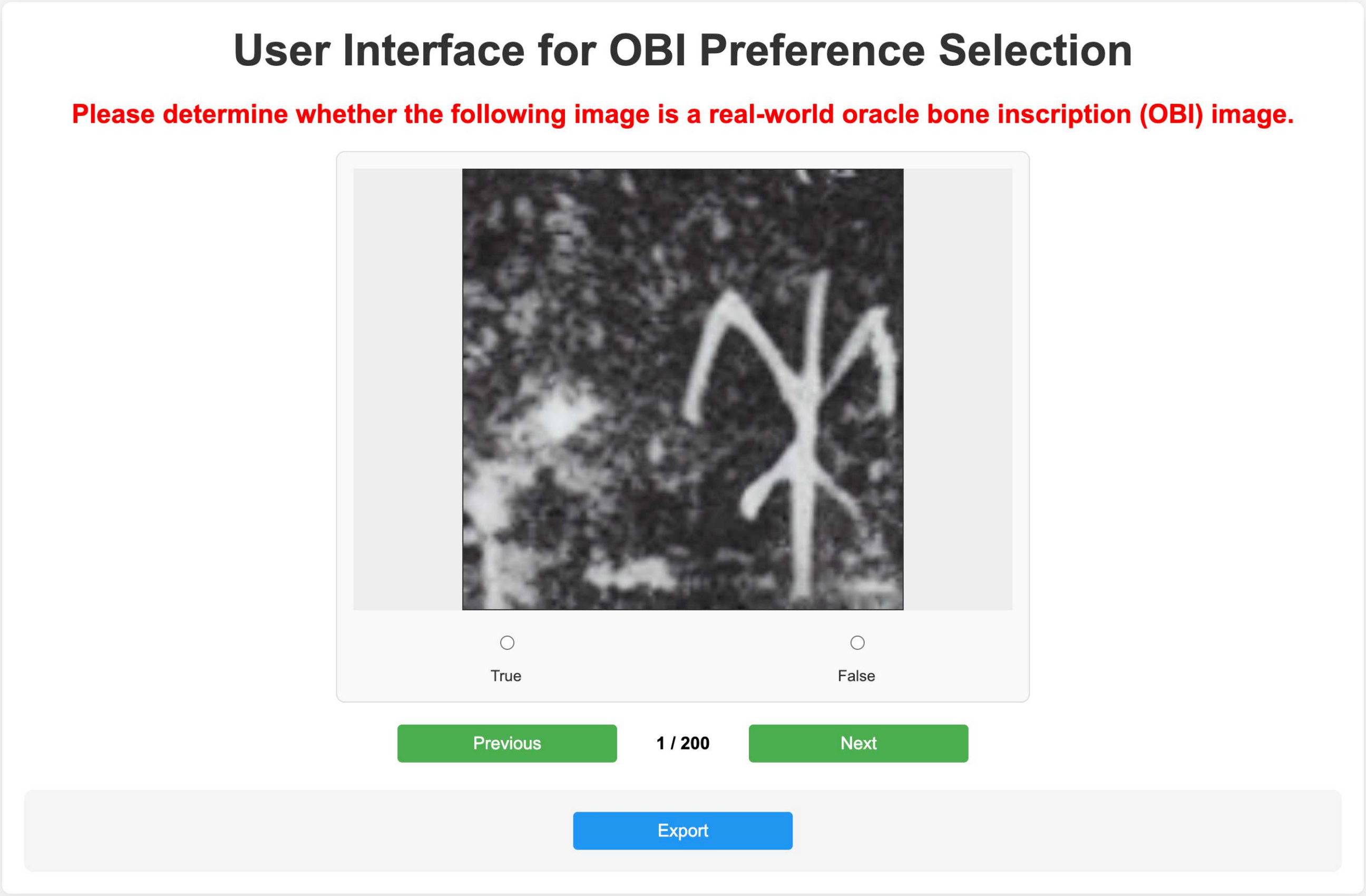}
    \caption{User interface of human evaluation in OBI preference selection task.}
    \label{fig:interface}
\end{figure}

We utilize HTML, CSS, and JavaScript to develop a human evaluation user interface for OBI. As shown in \cref{fig:interface}, the interface presents either a ground truth image or a pseudo image generated by our OBIDiff in each round. Participants can click the button below to determine whether the current image is a real-world OBI rubbing image or a generated image. The navigation within the evaluation workflow is facilitated by \textit{Previous} and \textit{Next} buttons. Once the evaluation is complete, the user can click the \textit{Export} button to generate a brief evaluation report. This report includes relevant user information and an assessment of the user's performance in the evaluation. We present the detailed results of the user preference study in \cref{tab:denoising}.
\begin{table*}[t]
    \caption{Detailed results of the user study.}
    \label{tab:denoising}
    \begin{tabular}{@{}ccccccccccccccccc@{}}
	\toprule
         & 1 & 2 & 3 & 4 & 5 & 6 & 7 & 8 & 9 & 10 & 11 & 12 & 13 & 14 & 15 & Average \\ 
        \midrule
        Precision (\%) & 0.56 & 0.55 & 0.50 & 0.53 & 0.49 & 0.48 & 0.52 & 0.54 & 0.58 & 0.47 & 0.54 & 0.58 & 0.54 & 0.44 & 0.49 & 0.52 \\
        Recall (\%) & 0.53 & 0.86 & 0.88 & 0.44 & 0.62 & 0.56 & 0.40 & 0.56 & 0.48 & 0.16 & 0.63 & 0.68 & 0.63 & 0.47 & 0.66 & 0.57 \\
        F1 Score & 0.54 & 0.67 & 0.64 & 0.48 & 0.55 & 0.52 & 0.45 & 0.55 & 0.53 & 0.24 & 0.58 & 0.63 & 0.58 & 0.46 & 0.56 & 0.53 \\
        Duration (s) & 1126 & 2501 & 3207 & 1568 & 946 & 823 & 1015 & 867 & 1952 & 483 & 485 & 497 & 595 & 3155 & 2256 & 1432 \\
        \bottomrule
    \end{tabular}
\end{table*}

\begin{figure*}[t]
    \centering
    \includegraphics[width=\linewidth]{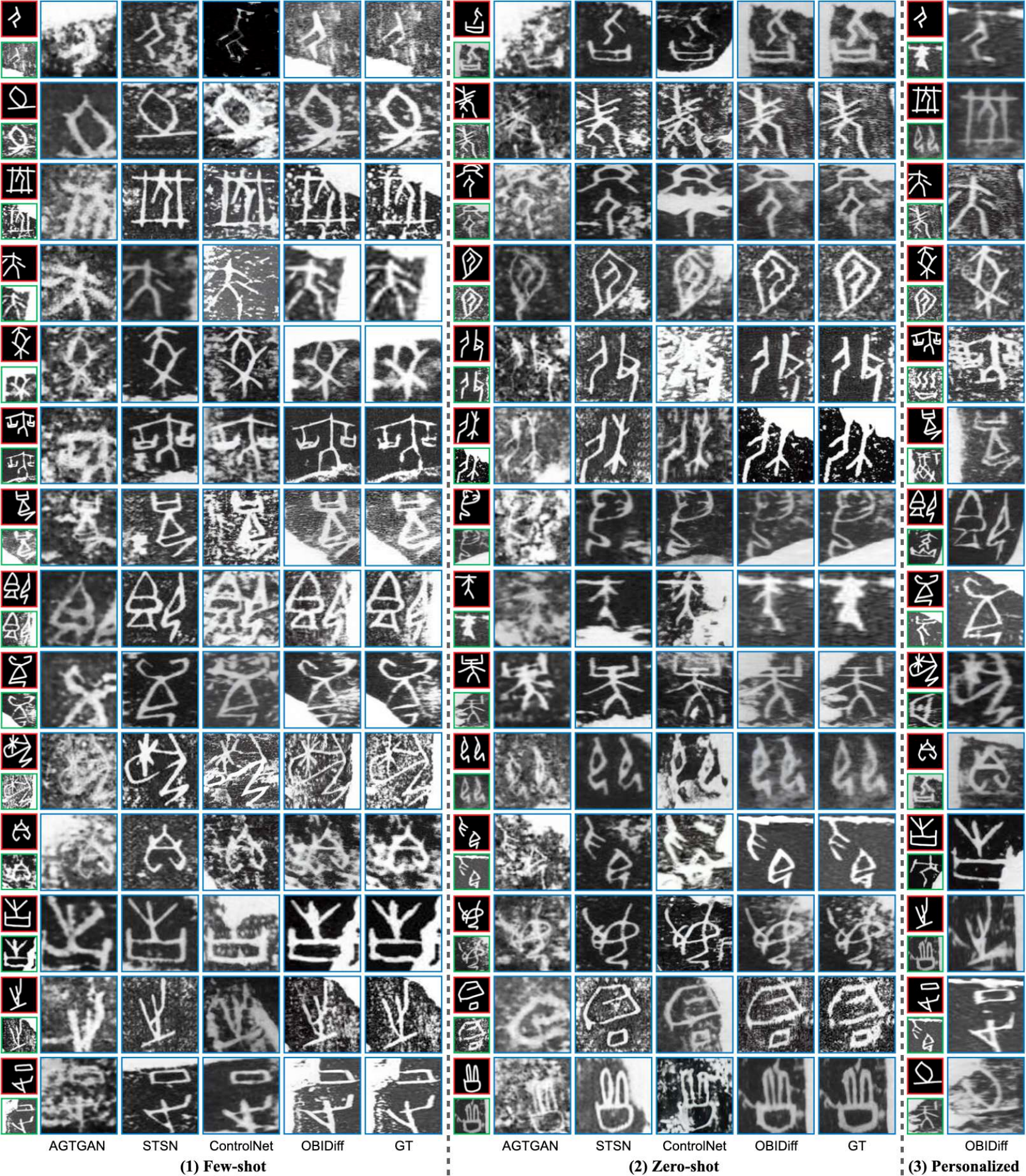}
    \caption{More generation results from our OBIDiff and other OBI generation methods.}
    \label{fig:appd_gen}
\end{figure*}

\begin{figure*}[t]
    \centering
    \includegraphics[width=\linewidth]{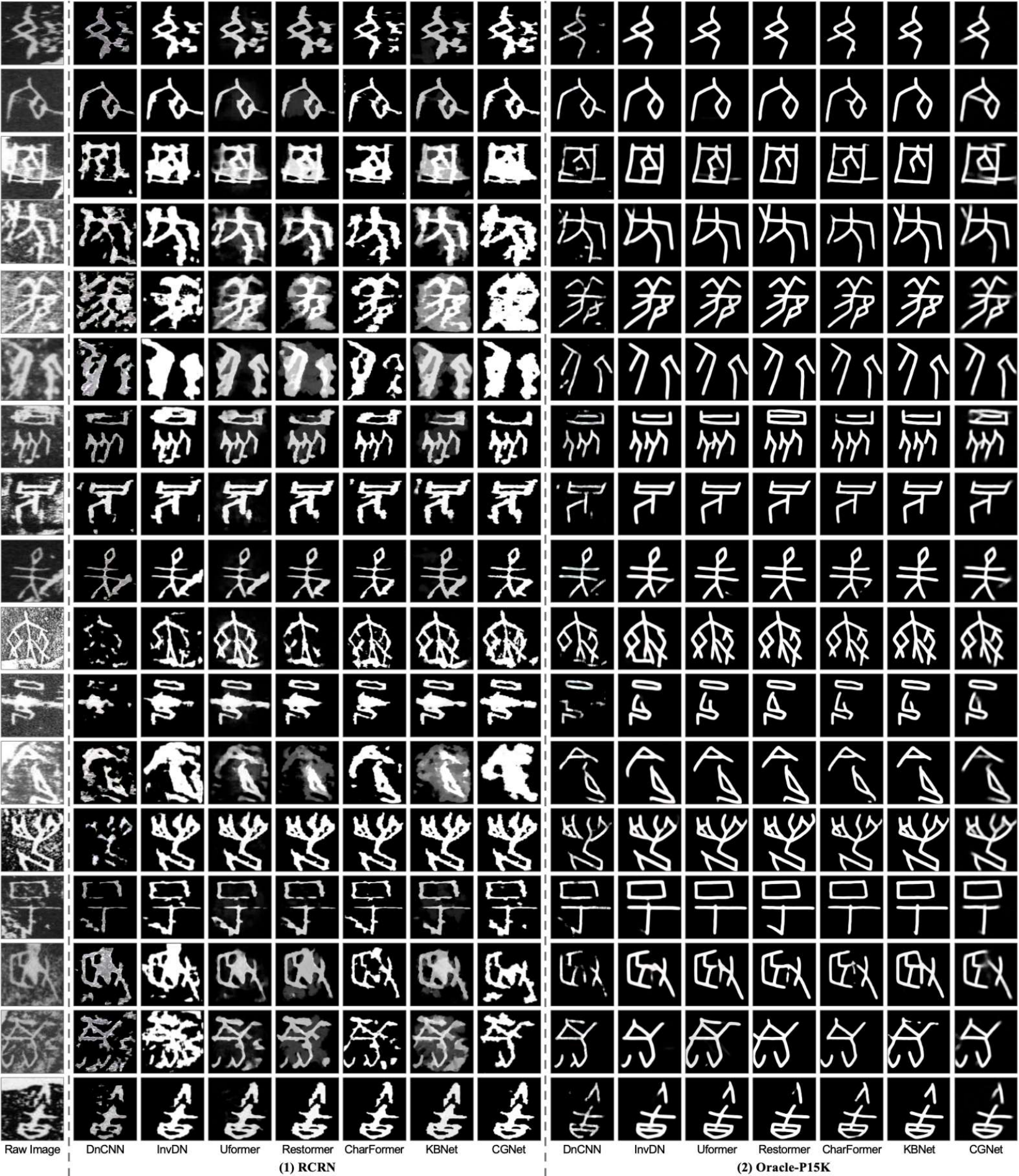}
    \caption{More denoising results from generic and OBI denoising baselines.}
    \label{fig:appd_den}
\end{figure*}

\end{document}